\documentclass[sigconf]{acmart}

\usepackage{booktabs} 
\usepackage[utf8]{inputenc}
\usepackage{listings}

\setcopyright{rightsretained}
\acmDOI{TODO}

\acmISBN{}

\acmConference[arXiv Preprint]{}
\copyrightyear{2017}

\renewcommand\footnotetextcopyrightpermission[1]{}

\begin{document}
\title{Fine-tuning deep CNN models on specific MS COCO categories}
\subtitle{DFKI library: py-faster-rcnn-ft}

\author{Daniel Sonntag}
\affiliation{
  \institution{German Research Center for Artificial Intelligence (DFKI)}
  \city{Saarbrücken}
  \country{Germany}
}
\email{daniel.sonntag@dfki.de}

\author{Michael Barz}
\affiliation{
  \institution{German Research Center for Artificial Intelligence (DFKI)}
  \city{Saarbrücken}
  \country{Germany}
}
\email{michael.barz@dfki.de}

\author{Jan Zacharias}
\affiliation{
  \institution{German Research Center for Artificial Intelligence (DFKI)}
  \city{Saarbrücken}
  \country{Germany}
}
\email{jan.zacharias@dfki.de}

\author{Sven Stauden}
\affiliation{
  \institution{German Research Center for Artificial Intelligence (DFKI)}
  \city{Saarbrücken}
  \country{Germany}
}
\email{sven.stauden@dfki.de}

\author{Vahid Rahmani}
\affiliation{
  \institution{German Research Center for Artificial Intelligence (DFKI)}
  \city{Saarbrücken}
  \country{Germany}
}
\email{vahid.rahmani@dfki.de}

\author{Áron Fóthi}
\affiliation{
  \institution{Faculty of Informatics\\ Eötvös Loránd University}
  \city{Budapest}
  \country{Hungary}
}
\email{aronfothi@gmail.com}

\author{András L{\H o}rincz}
\affiliation{
  \institution{Faculty of Informatics\\ Eötvös Loránd University}
  \city{Budapest}
  \country{Hungary}
}
\email{lorincz@inf.elte.hu}

\renewcommand{\shortauthors}{D. Sonntag et al.}

\begin{abstract}
Fine-tuning of a deep convolutional neural network (CNN) is often desired. This paper provides an overview of our publicly available {\bf py-faster-rcnn-ft} software library that can be used to fine-tune the VGG\_CNN\_M\_1024 model on custom subsets of the Microsoft Common Objects in Context (MS COCO) dataset. For example, we improved the procedure so that the user does not have to look for suitable image files in the dataset by hand which can then be used in the demo program. Our implementation randomly selects images that contain at least one object of the categories on which the model is fine-tuned. 
\end{abstract}


\maketitle

\section{Introduction}

Understanding of visual scenes is an important goal for many artificial intelligence applications where knowledge acquisition, anomaly detection and intelligent user interfaces are central parts. For example, human-robot interaction in Industry 4.0 factories \cite{Sonntag2017}; medical decision support \cite{ijcai2017-777} where scene understanding involves numerous tasks including localizing objects and images in 2D and 3D; contextual reasoning between objects and the precise 2D localization of objects \cite{pub8992}. Image recognition is a central part of the technical architectures of these applications. In order to provide state-of-the-art image classifiers, Chatfield et al. trained their VGG\_CNN\_M\_1024 model \cite{Chatfield2014} on the ImageNet ILSVRC-2012 dataset \cite{Russakovsky2015} that contains about 1.2M training images categorized into 1000 object classes. Girshick et al. showed that fine-tuning a model on the target dataset improves mean average precision \cite{Girshick2012}. Chatfield et al. could verify this result by fine-tuning their model on the PASCAL-2007 VOC~\cite{everingham2010pascal} dataset that contains only 2501 training images in 20 categories.

Our approach is a fork of \texttt{py-faster-rcnn}\footnote{https://github.com/rbgirshick/py-faster-rcnn} by Ren et al. \cite{Ren2015} which uses approximate joint training where the region proposal network is trained jointly with the FAST R-CNN network resulting in overall speed improvements. \texttt{py-faster-rcnn-\textbf{ft}} allows for a convenient \textbf{f}ine-\textbf{t}uning of the VGG\_ CNN\_M\_1024 model on specific object classes of the MS COCO dataset \cite{Lin2014}. We found it cumbersome that the user needed to enter these IDs in source code at several locations. Furthermore, the necessity of user inputs is error-prone and could cause incorrect states.
The software is freely available under the GPLv3 license\footnote{https://www.gnu.org/licenses/gpl-3.0.en.html} via our GitHub repository\footnote{https://github.com/DFKI-Interactive-Machine-Learning/py-faster-rcnn-ft}; developer feedback is very much appreciated. 

\section{Related Work and Background}
In a number of cases, fine-tuning of the network is desired. This can be due to the environment that may strongly influence the best scores \cite{lorincz2017towards}. Alternatively, if evaluation is followed by a consistence seeking procedure that exploits more than one network, then categorization can be fixed and new training samples can be produced as demonstrated in \cite{lorincz2017deep}. In such cases, retraining the network(s) concerns only a few samples, and fine-tuning is the desired way to go:

\begin{itemize}
\item R-CNN works with bounding boxes (BBs). Any BB has some part of the background. Background dependence can be strong and can easily mislead the categorization as shown in Figure~\ref{fig:demo}. In turn, different environments may need fine-tuning.

\item Faster R-CNN gains speed by sharing full image convolutional features with the detection network.

 The net result is that the time required by region proposals is almost negligible compared to the rest of the computations. This solution is of high interest as it can be seen as a special version of transfer learning.
\end{itemize}

\section{Implementation}
The original \texttt{py-faster-rcnn} software works with a model trained on a dataset containing image files and the corresponding annotations. The model is able to detect and classify objects in the input images for several classes. In our work we use the pre-trained model VGG\_CNN\_M\_1024 that is able to differentiate between 1000 classes relating to the ILSVRC-2012 challenge. We improved this original software as explained in the rest of this paper, in order to be able to fine-tune on specific image dataset categories.

\begin{figure}
 \centering
 \includegraphics[width=\linewidth]{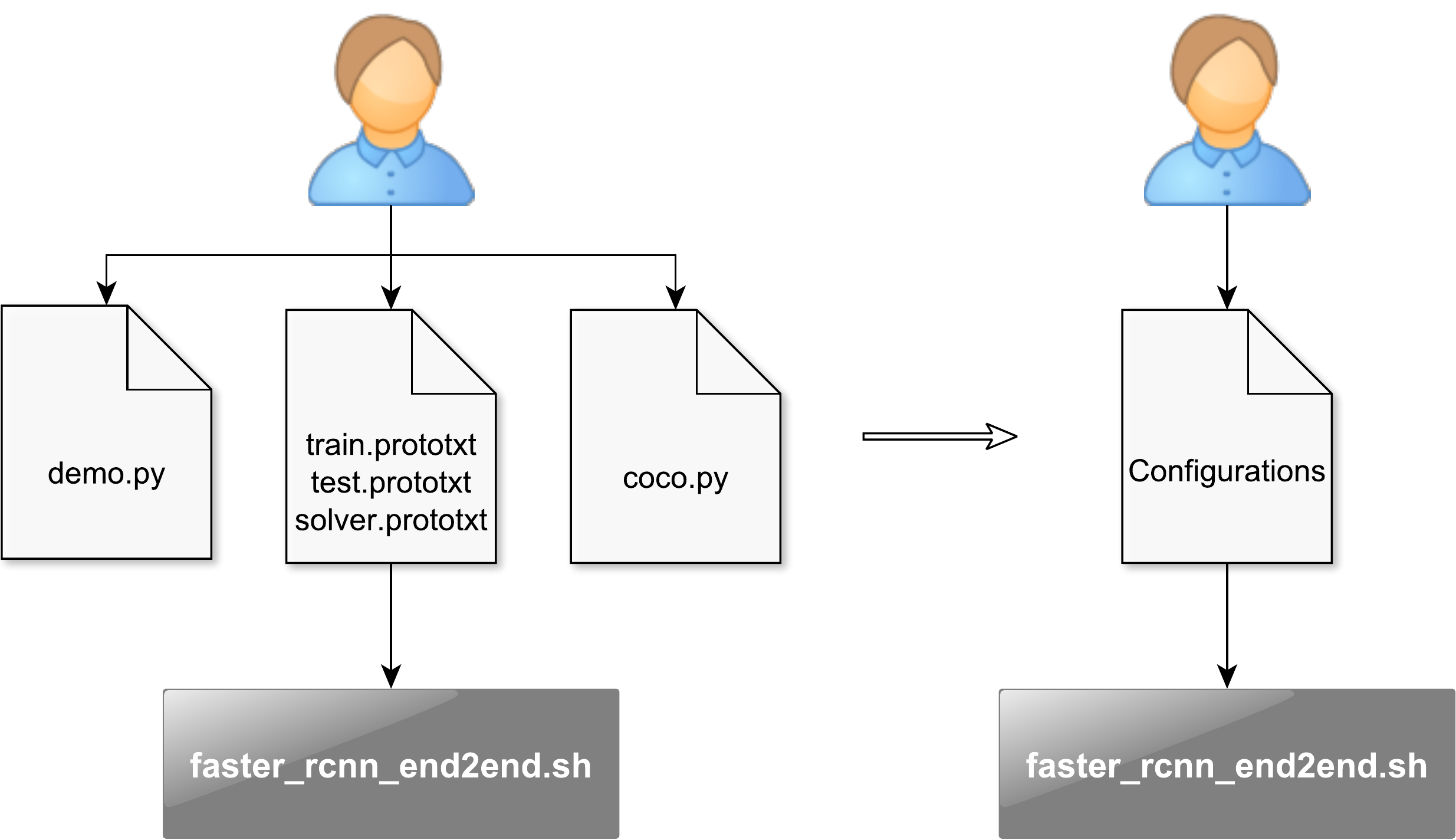}
 \caption{Transition of the setup workflow for training with the original py-faster-rcnn (left) and our fork (right)}
 \label{fig:setup}
\end{figure}

\subsection{Prototxt File Managing}
Fine-tuning a model for a subset of categories requires changes of some aspects in the architecture of the neural network which will be applied in the retraining process.
To perform fine-tuning, several parameters in the corresponding train.prototxt and test.prototxt files need to be continuously changed, depending on the chosen subset of classes the fine-tuning should perform on. Because the manual change of these parameters in the files often leads to errors, we implemented a python wrapper that is able to read and manipulate prototxt files. In our implementation, this wrapper performs the prototxt file manipulation automatically in the background (hyper-parameter selection).

\begin{figure*}
 \includegraphics[width=\linewidth]{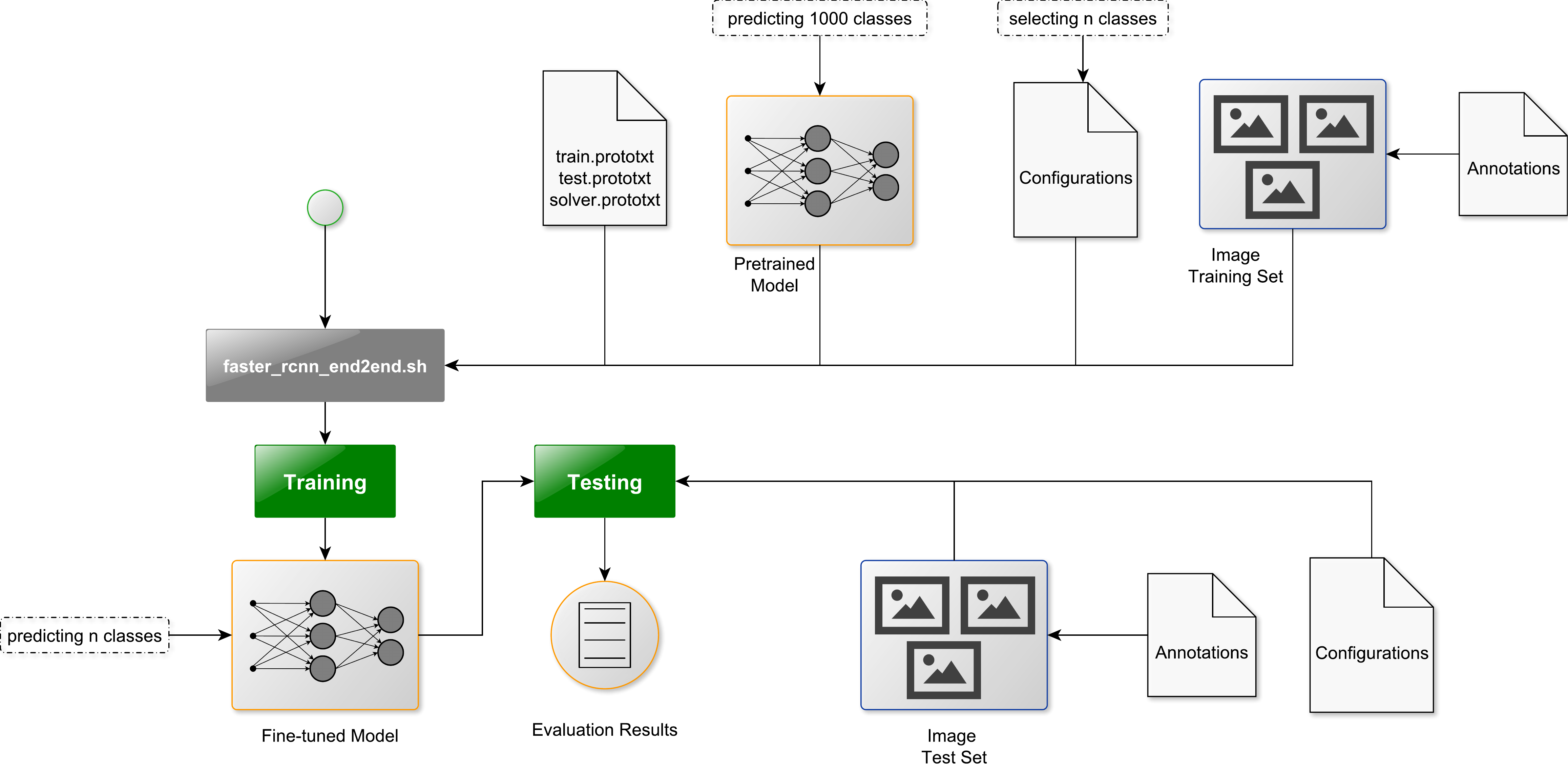}
 \caption{Training with \texttt{faster\_rcnn\_end2end.sh} process diagram}
 \label{fig:process}
\end{figure*}

\subsection{MS COCO Category IDs}
For the training process, the user has to decide on which categories the fine-tuning should be performed. As the MS COCO dataset manages its classes with so-called category IDs, we wrote a program that extracts all classes of the MS COCO 2014 dataset along with their IDs. This program is available at: \texttt{data/scripts/MSCOCO\_API\_categories.py} and helps the user to quickly decide on the correct category IDs.

\subsection{Config File Settings}
\label{s:config}
The dataset MS COCO works with category IDs. We found it cumbersome that the user needed to enter these IDs in the source code at several locations. Furthermore, the necessity of user inputs is error-prone and could cause incorrect states.

To mitigate this risk, we implemented the possibility for the user to enter the category IDs just once in an already existing configuration file (\texttt{experiments/cfgs/faster\_rcnn\_end2end.yml}) under the keyword \texttt{CAT\_IDS}. All necessary changes are done automatically by our implementation. This effectively reduces erroneous user inputs and simplifies the setup workflow (cf. Figure~\ref{fig:setup}).

\subsection{Demo Image Selection}
The demo program \texttt{tools/demo.py} is an important tool for visualizing the performance of a trained or fine-tuned model. In the original version, the user is able to choose images from the dataset manually and apply the model on them while predicted bounding box regions and labels are plotted as an overlay on the corresponding image. We improved this procedure so that the user does not have to manually search for suitable image files in the dataset which can subsequently be used in the demo program. Our implementation randomly selects images that contain at least one object of the categories on which the model is fine-tuned.

\subsection{Bugfixes}
As described in Section~\ref{s:config}, the user can fine-tune on a subset of classes by filling up CAT\_IDS in the configuration file. In the training stage, the software creates a list of all training samples containing annotation information. The original software does not distinguish between data samples of selected or unselected categories. This leads to the creation of many arrays with invalid data causing a lot of crashes and false results. In \texttt{lib/datasets/coco.py}, we filter invalid data and make sure that only samples from the chosen categories are used for the fine-tuning step.

\section{Usage}
To allow for a quick setup time and rapid results, we provide information regarding the installation of the software and how to fine-tune a model as well as verifying performance of the newly trained model. This is explained in the following subsections. A complete step-by-step installation instruction and usage listings are available in the README file\footnote{https://github.com/DFKI-Interactive-Machine-Learning/py-faster-rcnn-ft/blob/master/README.md} accompanying our software library.

\subsection{Requirements}
We suggest the installation of the Ubuntu 17.04 64-bit operating system\footnote{http://releases.ubuntu.com/zesty/} as this allows for an easy and quick installation with minimal compiling from addional sources. A computer with at least one GPU supporting CUDA\footnote{https://developer.nvidia.com/cuda-downloads} is required for the operation of \texttt{py-faster-rcnn-ft}. A single Nvidia GTX 1080 was found to be sufficient for the experiments with the VGG\_CNN\_M\_1024 model.

\subsection{Installation}
After installing Ubuntu 17.04, the proprietary Nvidia driver needs to be activated in the \textit{Additional Drivers} tab of the \textit{Software \& Updates} settings. \textit{Canonical Partners} should be ticked in the \textit{Other Software} tab as well.

\noindent
\texttt{py-faster-rcnn-ft} requires a lot of additional software on a fresh installation, however most of it can be easily installed by using the \texttt{apt}\footnote{https://wiki.debian.org/Apt} command line tool, a root shell is required to be able to perform the installation of the following packages: \texttt{\small{python-pip python-opencv libboost-dev-all libgoogle-glog-dev libgflags-dev libsnappy-dev libatlas-dev libatlas4-base \\ libatlas-base-dev libhdf5-serial-dev liblmdb-dev libleveldb-dev libopencv-dev g++-5 nvidia-cuda-toolkit cython python-numpy python-setuptools python-protobuf python-skimage python-tk \\ python-yaml}}.

\noindent
The python package \texttt{easydict} can then be installed via \texttt{pip2}. Furthermore \texttt{cuDNN}\footnote{https://developer.nvidia.com/cudnn} needs to be installed, this requires a registration with Nvidia. Note that the distribution version of \texttt{protobuf} can not be used: for compatibility with CUDA/cuDNN this needs to be cloned from the repository\footnote{https://github.com/google/protobuf.git} and compiled by the user with \texttt{gcc-5}.

The compilation of the Cython modules, Caffe and pycaffe, as provided with our repository, is straightforward. Trained models and datasets need to be downloaded separately as these big files are not suitable for a GitHub repository:
\begin{lstlisting}
git clone https://github.com/DFKI-Interactive-Machine-Learning/py-faster-rcnn-ft.git
cd py-faster-rcnn-ft/lib
make -j8 &&
cd ../caffe-fast-rcnn && make -j8 && make pycaffe &&
cd .. && data/scripts/fetch_imagenet_models.sh
\end{lstlisting}

\subsection{Demo}
\begin{figure}
\includegraphics[width=\linewidth]{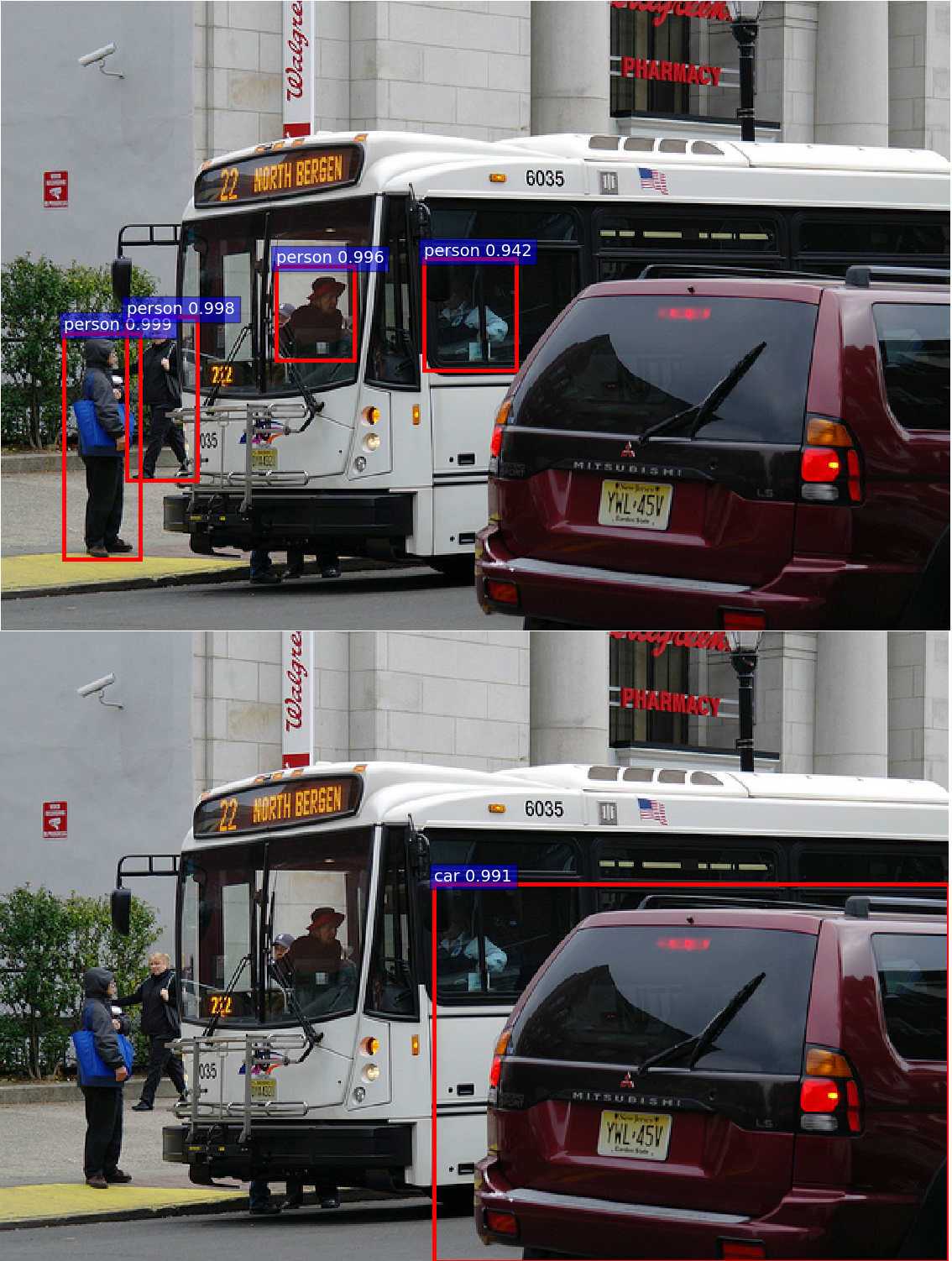}
\caption{\texttt{demo.py} ex. output with classification results and AP for categories person (top) and car (bottom)}
\label{fig:demo}
\end{figure}

The successful installation can be verified by running the provided \texttt{demo.py} in the \texttt{tools} subfolder. This demo uses the MS COCO dataset which needs to be downloaded along with an already fine-tuned model with the classes \textit{person} and \textit{car}. The demo program randomly selects images and tries to classify the content and display the result (cf. Figure~\ref{fig:demo}).

\subsection{Fine-tuning on Specific Classes}
\label{s:fine-tuning}
In order to fine-tune the VGG\_CNN\_M\_1024 model on specific categories of the MS COCO dataset, the respective category IDs are specified in the \texttt{experiments/cfgs/faster\_rcnn\_end2end.yml} configuration file. Possible IDs can be listed by running \texttt{MSCOCO\_API \_categories.py} inside the \texttt{data} folder.

The fine-tuning process (cf. Figure~\ref{fig:process}) is started by:
\begin{lstlisting}
./faster_rcnn_end2end.sh 0 VGG_CNN_M_1024 coco
\end{lstlisting}
The first argument denotes the GPU ID to use while 0 is the first GPU in the system. The training will take about 12 hours with the default and recommended iteration setting of 490000 and one Nvidia GTX 1080 graphics card. Following the training, an automatic test run is started with the new model and the MS COCO minival2014 validation test dataset which is comprised of 5000 images. These defaults can be changed by editing the \texttt{faster\_rcnn\_end2end.sh} script.

If the demo program should use the newly trained model, the caffe-model must be moved from the \texttt{output} directory (e.g. \texttt{output /faster\_rcnn\_end2end/coco\_2014\_train/vgg\_cnn\_m\_1024\_}\newline\texttt{faster\_rcnn\_iter\_49000.caffemodel}) to \texttt{data/faster\_rcnn}\newline\texttt{\_models/<yourmodel.caffemodel>} and the \texttt{NET} dict in \texttt{demo.py} updated accordingly.

\subsection{Testing a Fine-tuned Model}
Testing of fine-tuned models can be started by using the \texttt{test\_net.py} program in the \texttt{tools} directory:
\begin{lstlisting}
tools/test_net.py --gpu 0 --def models/coco/VGG_CNN_M_1024/faster_rcnn_end2end/test.prototxt --net data/faster_rcnn_models/vgg_cnn_m_1024_faster_rcnn_iter_490000.caffemodel --imdb coco_2014_val --cfg experiments/cfgs/faster_rcnn_end2end.yml
\end{lstlisting}
This example tests the specified caffe-model against the \texttt{coco\_2014\_val} validation image dateset consisting of 40504 images, hence the testing will take longer. In case the smaller validation dataset should be used, \texttt{--imdb coco\_2014\_minival} can be specified.

Note that the prototxt file will be updated automatically with the category IDs as specified in the config file analogously to the fine-tuning (cf. Section~\ref{s:fine-tuning}).

\subsection{Informal Evaluation}
The provided caffemodel\footnote{https://www.dfki.de/$\sim$jan/vgg\_cnn\_m\_1024\_faster\_rcnn\_iter\_490000.caffemodel} was fine-tuned on the categories person and car. When tested with the minival2014 dataset, the AP @ IoU=[0.50,0.95] for person is 29.4\% and 15.6\% for car. A model trained on all 80 MS COCO categories results in 26.2\% for person and 11.2\% for car. These early results indicate an improvement of average precision when fine-tuning is performed on a subset of the target dataset.

\section{Future Work} 
\texttt{py-faster-rcnn-ft} is limited to the MS COCO dataset, we would like to be able to fine-tune by using other datasets that could be automatically generated using a pre-trained model, i.e., we could use output from Google Images\footnote{https://images.google.com/} for new dataset instances but verify them beforehand. NLP techniques could be used to search for relevant images based on the category.

\bibliographystyle{ACM-Reference-Format}
\bibliography{paper}

\end{document}